\documentclass[sn-mathphys]{sn-jnl}

\usepackage{graphics} 
\usepackage{epsfig} 
\usepackage{mathptmx} 
\usepackage{times} 
\usepackage{amsmath} 
\usepackage{amssymb}  
\usepackage{rotating}
\usepackage{makecell}
\usepackage{multirow}
\usepackage{amsmath}
\usepackage{amssymb}
\usepackage{diagbox}
\usepackage{subfigure}
\usepackage{url}

\jyear{2022}%

\theoremstyle{thmstyleone}%
\theoremstyle{thmstyletwo}%

\theoremstyle{thmstylethree}%

\begin{document}

\title{Point Cloud Learning with Transformer}


\author[1]{\fnm{Qi} \sur{Zhong}}\email{cuihuazq@email.swu.edu.cn}
\equalcont{These authors contributed equally to this work.}

\author*[1]{\fnm{Xian-Feng} \sur{Han}}\email{xianfenghan@swu.edu.cn}
\equalcont{These authors contributed equally to this work.}

\affil*[1]{\orgdiv{College of Computer and Information Science}, \orgname{Southwest University}, \orgaddress{\street{Tiansheng Road}, \city{Chongqing}, \postcode{400715}, \country{China}}}


\abstract{ Remarkable performance from Transformer networks in Natural Language Processing promote the development of these models in dealing with computer vision tasks such as image recognition and segmentation. In this paper, we introduce a novel framework, called Multi-level Multi-scale Point Transformer (MLMSPT) that works directly on the irregular point clouds for representation learning. Specifically, a point pyramid transformer is investigated to model features with diverse resolutions or scales we defined, followed by a multi-level transformer module to aggregate contextual information from different levels of each scale and enhance their interactions. While a multi-scale transformer module is designed to capture the dependencies among representations across different scales. Extensive evaluation on public benchmark datasets demonstrate the effectiveness and the competitive performance of our methods on 3D shape classification, segmentation tasks. }




\maketitle

\section{Introduction}

Recently, point cloud analysis has been drawing more and more attention, since point cloud, becoming a preferred representation for tasks of classification and segmentation, can provide much richer geometric as well as  photometric information in comparison with 2D images. Specifically, 
the outstanding success of deep learning strategies further make the task of 3D point cloud analysis achieve remarkable advancements in a diverse range of applications, such as autonomous driving \cite{guo2020deep}\cite{hu2020randla}, robotics \cite{nezhadarya2020adaptive}\cite{wang2019deep}, virtual/augmented reality \cite{gojcic2020learning}\cite{jiang2020end}. However, effective and efficient feature learning from point clouds is still a challenge problem due to the irregular, unordered and sparse nature of point clouds. 

To tackle such crucial challenges, many state-of-the-arts works focus on transforming the unstructured point cloud into either voxel grids \cite{maturana2015voxnet} or multi-view images \cite{su2015multi}. Although impressive progress has been made, these types of regular representation inevitably give rise to loss of underlying geometric information during transformation, as well as high computation cost and memory consumption. The appearance of point-wise methods, such as PointNet \cite{qi2017pointnet}, has revolutionized point cloud learning. These approaches directly process the raw point clouds by adopting shared Multi-Layer Perceptrons (MLP) \cite{qi2017pointnet++} or defining convolutional kernels \cite{thomas2019kpconv} or constructing graph \cite{lei2020spherical}.However, most existing approaches may not be effective enough to learn context-dependent representation for point clouds.

\begin{figure*}
\centering
\includegraphics[width=5in]{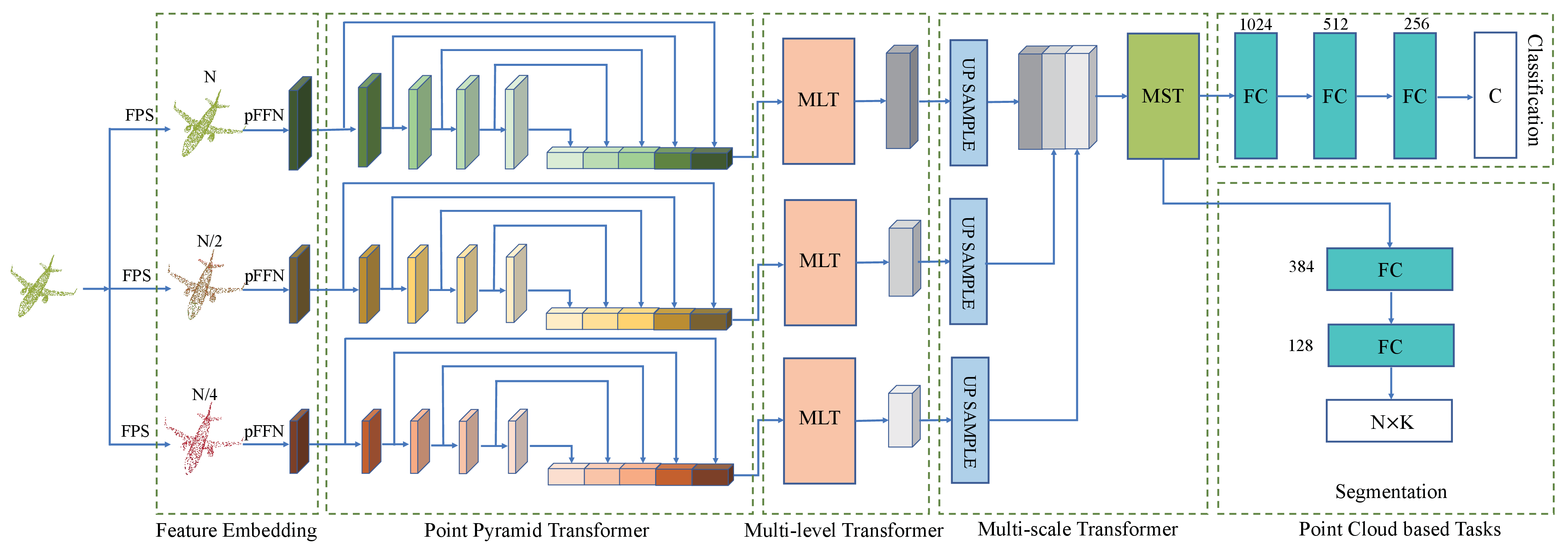}
\caption{The overall architecture of Multi-level Multi-scale Point Transformer model for point cloud analysis. The network mainly contains three key components: a Point Pyramid Transformer encoding pointwise features with three different scales or resolutions; a Multi-level Transformer and a Multi-scale Transformer modeling cross-level and cross-scale context dependencies representation. pFFN represents point Feature Forward Network.}
\label{fig_framework}
\end{figure*}

In this work, we propose a novel point-based transformer architecture, named Multi-level multi-scale transformer (MLMST) following the tremendous achievement by transformer models in the fields of Natural Language Processing (NLP) and 2D Computer Vision. Actually, transformer provide an idea strategy to model relationships between points, since it is permutation invariant and highly expressive as convolution.    
As shown in Figure \ref{fig_framework}, our MLMST mainly consists of three carefully-designed modules: (1) a point pyramid transformer (PPT), capturing context information from different resolution or receptive fields. (2) a multi-level transformer (MLT), learning the cross-level feature interaction to further aggregate geometric and semantic information and (3) a multi-scale transformer (MST), modeling the context interaction across different scales to improve the expressive capability. Based on these three core modules, we can report that our MLMST is able to better capture the long-range contextual dependencies from different levels and scales in an end-to-end manner. 

We evaluate the effectiveness and representation capability of our network on several public benchmark datasets (e.g., ModelNet \cite{wu20153d}, ShapNet \cite{yi2016scalable}) for 3D point cloud classification and segmentation tasks. Extensive results show that MLMST can achieve highly striking performance comparable to state-of-the-art methods.

In summary, the main contributions of our work are as follows:
\begin{itemize}
    \item  We design a novel point pyramid transformer, a multi-level transformer and a multi-scale transformer to capture the cross-level and cross-scale context-aware feature interaction to improve the discriminative power of learned representation.
    \item  Based on these three modules, we construct an end-to-end network architecture, named Multi-level Multi-scale Point Transformer (MLMSPT), taking the unstructured point clouds as input for highly effective geometric and semantic feature learning.
    \item Extensive experiments on challenging benchmarks demonstrate that our MLMSPT model presents state-of-the-art performance on the tasks of 3D object classification, as well as segmentation based on point clouds.
\end{itemize}

\section{Related Work}

\subsection{Deep learning on Point Clouds}
\label{relatedwork}
Recently, increasing attention has been paid to design deep neural networks for 3D point cloud learning, which achieve the state-of-the-art performance in the applications of 3D object classification \cite{fujiwara2020neural}, part segmentation \cite{lin2020convolution} as well as semantic segmentation \cite{zhang2020fusion}. In this section, we provide a brief review of these present approaches that specifically can be categorized into four groups:  

\textbf{Voxel based Methods} \cite{le2018pointgrid} attempt to voxelize the unstructured point clouds into regular volumetric grid structure, so that the standard 3D convolutional neural networks can be directly applied similarity as the image to learn descriptors. For example, VoxNet \cite{maturana2015voxnet} is a milestone towards real 3D learning. However, these approaches have difficulty in capturing high resolution or fine-grained features due to the sparsity, loss of geometric information during rasterization, as well as expensive memory and computational consumption \cite{nezhadarya2020adaptive}. Different efforts later have been made to alleviate this problem. OctNet \cite{riegler2017octnet} represents the 
point clouds using a hybrid grid-octree structure which is applicable to high resolution inputs of size $256\times256\times256$. And Kd-tree \cite{klokov2017escape} is another structure that can also be utilized to provide improved grid resolution.

\textbf{Multi-view based Methods} \cite{su2015multi} project the raw 3D point clouds into a collection of 2D images rendered from different viewpoints, followed by image-wise feature extraction with well-designed 2D convolutional neural networks. Then fusing these features forms the final output representation for various analyses. Although remarkable performance is achieved, this kind of approaches suffers from information loss during the projection process \cite{you2020pointwise} and becomes time-consuming when dealing with sparse point clouds. On the other hand, it is difficult to determine the appropriate number of views for modeling the underlying geometric structure \cite{zhang2020shape}.

\textbf{Point Cloud based Methods} directly manipulate unstructured and unordered point clouds, and take the 3D coordinates or/and RGB or/and normal as initial input. As the pioneering work, the emergence of PointNet \cite{qi2017pointnet} can be considered as a milestone in the domain of learning point clouds, which guides the development of pointwise MLP methods. This series of works usually utilize shared MLP to process each point individually to perform feature extraction. However, their performance is limited since they fail to capture local spatial relationships in the data  \cite{you2020pointwise}\cite{yang2020pbp}. Recent approaches begin to concentrate on defining effective convolution kernels for points. KPConv \cite{thomas2019kpconv} defines the point convolution using any number of kernel points with filter weights on each point, which gives more fleibgility and is invariant to point order. FPConv \cite{lin2020fpconv} proposes a surface-style convolution for point cloud analysis by learning local flattening, which can be treated as a complementary to the volumetric-style convolution.

\textbf{Graph based Methods} lead to a new trend of irregular data processing, which represent the point cloud as graph to model the local geometric information among points \cite{te2018rgcnn}. ECC \cite{simonovsky2017dynamic} and DGCNN \cite{wang2018dynamic} propose different edge-dependent convolution operations to aggregate neighboring features spatially. GAC \cite{wang2019graph} define the filter kernel using the learned attentional weights assigned to neighboring points, which is heleful for semantic segmentation.3D-GCN \cite{lin2020convolution} introduce a well-designed deformable 3D kernel to guarantee scale invariance, and a 3D graph max-pooling operation to summarize cross-scale features. SPH3D-GCN \cite{lei2020spherical} proposes a separable spherical convolutional kernel for graph neural networks. This network achieves highly competitive performance on the standard benchmarks.

\subsection{Transformer in Computer Vision}
The Transformer networks can be perceived as a significant breakthrough in Natural Language Processing (NLP), whose success is mainly attributed to the self-attention mechanisms which can model long-range information and dependencies in the input data \cite{khan2021transformers}. Recently, many architectures begin to take transformer and self-attention into consideration for computer vision task. Image GPT \cite{chen2020generative} is the first work to investigate the Transformer for learning image representation in an unsupervised fashion. ViT \cite{dosovitskiy2020image} applies the original Transformer to image patches instead of pixels for image classification task, which can achieve the state-of-the-art performance with less computational resources consumption. DETR \cite{carion2020end} performs object detection from the set prediction point of view using a transformer encoder-decoder architecture. The advantage of DETR is that it doesnot require the hand-designed modules (e.g., non-maximal suppression ) usually used in the previous frameworks. 

Inspired by the fundamental mechanism of Transformer in NLP and Computer Vision, we aim to model the cross-level and cross-scale feature dependencies to obtain fine-grained performance for point cloud based tasks with our well-designed transformer modules.

\section{Point Cloud Representation with Transformer} 
As illustrated in Figure \ref{fig_framework}, given an input point cloud of N points $P \in R^{N \times C}$, where C represents the dimension of pointwise properties. Here we aim at modeling cross-level cross-scale interaction to discrimnatively boost the expressive capability of learned representations. Farthest point sampling (FPS) is initially performed to obtain three point clouds with different resolutions, followed by potential feature learning with feature embedding module. Then, we extract hierarchically multi-scale representations via our Point Pyramid Transformer (PPT). For each path of PPT module, Multi-level Transformer (MLT) consumes the concatenation of features from different levels to capture point cross-level representation correlation or interaction. Finally, through a Multi-scale Transformer (MST), we relate the point feature across different resolutions to learn a discriminative representation. 

\subsection{Point Pyramid Transformer}
\begin{figure}
\centering
\includegraphics[width=3in]{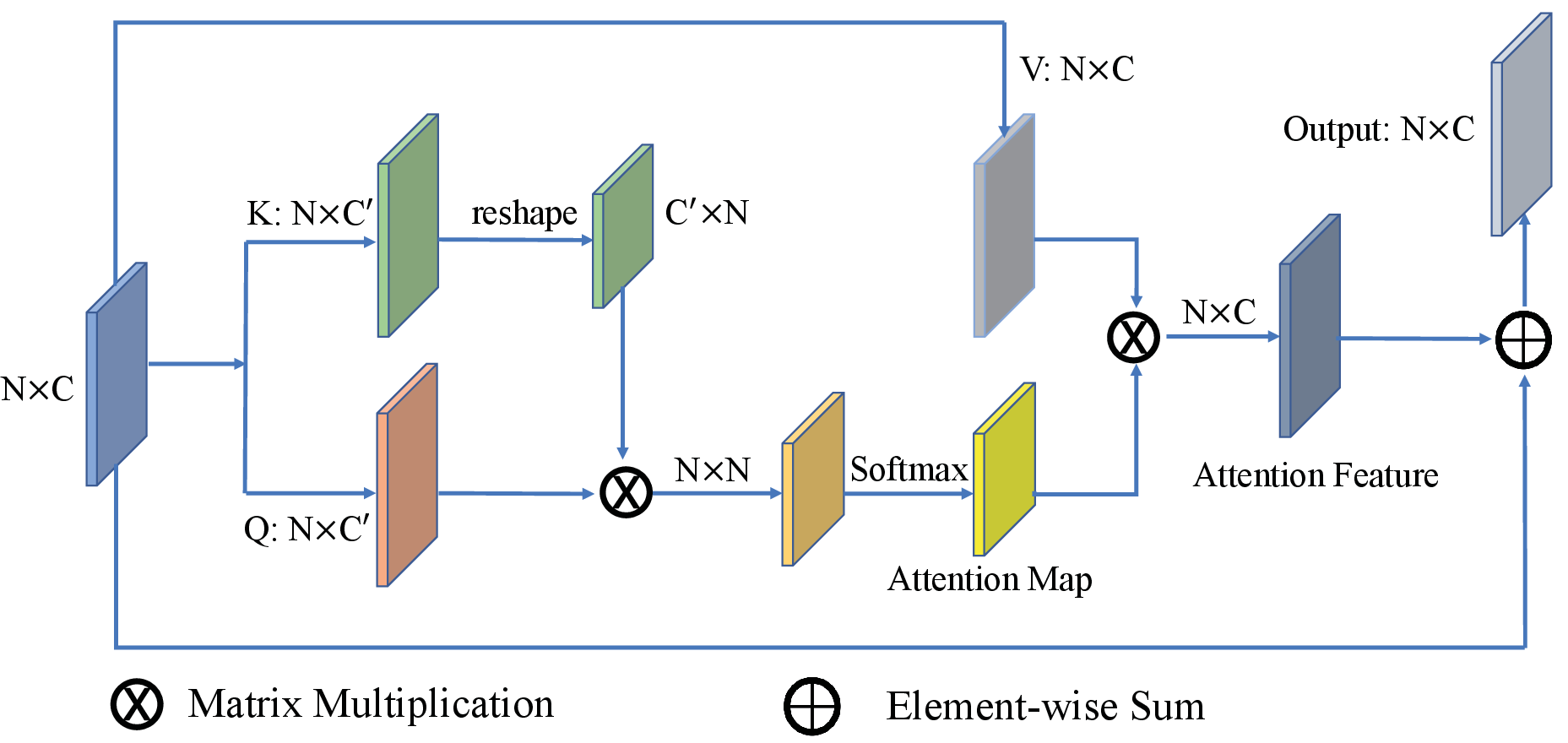}
\caption{Architecture of point self attention mechanism used in Point Pyramid Transformer. }
\label{fig_attention}
\end{figure}
Empirically, we can argue that different resolutions actually corresponding to different scales or having different size of receptive fields during feature extraction using the same operator. Therefore, in order to model a hierarchical semantic or contextual information with different scales for point cloud, we introduce a point pyramid transformer module. The farthest point sampling (FPS) operations are progressively performed on input point cloud $P$ to get three point clouds $P_{1}$, $P_{2}$, $P_{3}$ with different resolutions $N_{1} = N$, $N_{2} = N/2$ and $N_{3} = N/4$, respectively, followed by generating an initial pointwise feature map pyramid $\mathcal{P} = \{ F_{i}^{0} \in R^{N_{i} \times D}, i=1, 2, 3\} $ via feature embedding block using pointwise feedforward network. . 

The Point Pyramid Transformer (PPT) module takes $\mathcal{P}$ as input, where each branch independently maps corresponding scale feature maps into latent representation $\mathcal{F}_{PPT}^{i}\in R^{N_{i} \times D^{''}}, i = 1, 2, 3$. 
Here, since self-attention mechanism, as the core of Transformer models, is permutation invariant and can model long-range context dependencies, the essential building block of PPT, therefore, will based on point self-attention mechanism (PSA) as shown in Figure \ref{fig_attention}. Formally, we formulate the attention operator as follows:
\begin{equation}
    \mathcal{F}_{i}^{l} = \mathcal{A}(F_{i}^{l}) = \sigma (\Psi(F_{i}^{l})\Phi(F_{i}^{l})^{T}/\sqrt{D})\Delta(F_{i}^{l}) + F_{i}^{l}
\end{equation}
where $i=1, 2, 3$ denotes the $i$th scale or resolution, $l=0, 1, 2, 3, 4$ represents the $l$th layer. $\Psi(\bullet)$,  $\Phi(\bullet)$ and $\Delta(\bullet)$ are linear point transformation used in our paper. 
\begin{equation}
\Psi(F_{i}^{l}) = F_{i}^{l}W^{q} \in R^{N_{i} \times D^{'}}   
\end{equation}
\begin{equation}
\Phi(F_{i}^{l}) = F_{i}^{l}W^{k} \in R^{N_{i} \times D^{'}}   
\end{equation}
\begin{equation}
\Delta(F_{i}^{l}) = F_{i}^{l}W^{v} \in R^{N_{i} \times D}    
\end{equation}
$W^{q}, W^{k} \in R^{D \times D^{'}}, W^{v}\in R^{D \times D}$ define the learnable weight parameters. $\sigma$ adopts the softmax function to normalize the weights.

Subsequently, from top to bottom in our PPT, each path is constructed by stacking four sequential PSA modules to obtain pointwise feature representation from certain resolution of point cloud. Finally, to fully investigate cross-level information, feature maps in different levels together with the initial input are concatenated to generate the output of PPT module $\mathcal{F}_{ppt}^{i}$.
\begin{equation}
    \mathcal{F}_{ppt}^{i} = concat(F_{i}^{0}, \mathcal{F}_{i}^{1}, \mathcal{F}_{i}^{2}, \mathcal{F}_{i}^{3}, \mathcal{F}_{i}^{4}), i=1, 2, 3
\end{equation}

Actually, we can modify the number of resolution or scale branches, the stacked PSA modules or levels, and the size of input feature maps according to specific applications. 

\subsection{Multi-level Transformer}
Theoretically, aggregating information contained in different levels can boost the expressive power of pointwise features \cite{zhang2020feature}. In order to take fully advantage of multi-level context and model the long-range dependencies or interaction across these levels, we introduce Multi-level Transformer (MLT) module based on multi-head self-attention mechanism, which consists of three independently parallel MLT corresponding to each scale or resolution path.

Our MLT operator consumes $\mathcal{F}_{PPT}^{i}$ to encode much richer relationships amongst points, which is formally defined as: 

\begin{equation}
    \mathcal{F}_{MLT}^{i} = \mathcal{MA}(\mathcal{F}_{PPT}^{i})= concat(\mathcal{A}^{1}, \mathcal{A}^{2}, ..., \mathcal{A}^{M}) + \mathcal{F}_{PPT}^{i}
\end{equation}
where 
\begin{equation}
    \mathcal{A}^{m}(\mathcal{F}_{PPT}^{i}) = \sigma(Q^{i}_{m}(K^{i}_{m})^{T}/\sqrt{D^{''}/M}) V^{i}_{m}
\end{equation}
\begin{equation}
    Q^{i}_{m} = \mathcal{F}_{PPT}^{i}W_{Q_{m}}^{i}
\end{equation}
\begin{equation}
 K^{i}_{m} = \mathcal{F}_{PPT}^{i}W_{K_{m}}^{i}
\end{equation}
\begin{equation}
 V^{i}_{m} = \mathcal{F}_{PPT}^{i}W_{V_{m}}^{i}
\end{equation}

Here, $M$ is the number of heads. $m$ indicates the $m$th head. $W_{Q_{m}}^{i} \in R^{D^{''}\times d_{q}}$, $W_{K_{m}}^{i} \in R^{D^{''}\times d_{k}}$, $W_{V_{m}}^{i} \in R^{D^{''}\times d_{v}}$ are learnable weight matrices. We set $d_{q} = d_{k} = d_{v} = D^{''} / M$. 

Lastly, these three individual MLT map the level-concatenated features with three resolutions from PPT into three independent shape-semantic aware representation $\{\mathcal{F}_{MLT}^{i}, i = 1, 2, 3\}$.    

\subsection{Multi-scale Transformer}
Generally, point features from different scales or resolutions corresponding to different contextual or semantic information \cite{huang2020pf}. To enhance interaction among low-, mid- and high-resolution, we also adopt multi-head self-attention mechanism to construct our multi-scale Transformer (MST) module and generates relations across different scales.

The obtained cross-level interacted feature maps $\mathcal{F}^{i}_{MLT}$ from MLT are fed into our MST model. We first sample the maps of these three different scales directly up to the same size as the original input to our network via interpolation operation used in PointNet++ \cite{qi2017pointnet++}. Then we concatenate them together to encode multi-scale information.
\begin{equation}
    \mathcal{F}^{i}_{up} = Up(\mathcal{F}^{i}_{MLT}), i = 1, 2, 3
\end{equation}
\begin{equation}
    \mathcal{F}_{cat} = concat(\mathcal{F}^{1}_{up}, \mathcal{F}^{2}_{up}, \mathcal{F}^{3}_{up})
\end{equation}

Subsequently, the multi-scale transformer operation is performed on the $\mathcal{F}_{cat}$. With the similar multi-head self-attention strategy as Equ.(6), the formulation of MST module is defined as: 

\begin{equation}
     \mathcal{F}_{MST} = \mathcal{MA}(\mathcal{F}_{cat})
\end{equation}

By integrating Multi-scale Transformer module, we can  further boost the ability of our network to learn a discriminative representation for each point with semantically and geometrically enhanced information.

\section{Experiments}
\label{experiment}
In order to evaluate the performance of our proposed Multi-level Multi-scale Transformer model, we conduct extensive experiments for the problems of 3D point cloud classification, segmentation and provide comparison with the state-of-the-art approaches. Here, we adopt Pytorch framework to implement our Transformer architecture on NVIDIA Titan RTX with 24G memory. And Adam optimizer and step LR learning decay scheduler are used to train all the models. The following expands on the discussion of experiments and results.

\subsection{Point Cloud Classification}
\subsubsection{Datasets} The classification task is performed on ModelNet40 benchmark datasets. ModelNet40 consists of 12,311 CAD models from 40 categories, in which 9,843 instances are selected for training and 2,468 shapes are utilized for testing. Following PointNet\cite{qi2017pointnet}, 1,024 points are uniformly sampled from each object model. During training, we perform operations, including random point dropout, random scaling in [0.6, 1.55] and random shifting in [-0.2, 0.2] on input point clouds for data augmentation. We train the classification network for 250 epochs using an initial learning rate 0.01, which is dynamically adjusted using cosine annealing strategy. The batch size is set to 32.

\subsubsection{Performance Comparison} Table \ref{tab:cls} quantitatively reports the experimental comparisons with several state-of-the-art methods. Our MLMST directly operates on the raw xyz coordinates of only 1,024 points to yield these results. Specifically, on ModelNet40, our model achieves the best performance with a superior accuracy $93.2\%$, outperforming the voxel grid-input, multiple views-input and almost all state-of-the-art point-input methods. These competitive results convincingly demonstrate the effectiveness of our MLMST.

\begin{table}[]
\centering
\caption{3D Object Classification results on ModelNet40 dataset.}
\begin{tabular}{cccc} \hline
Method & Representation   & Input Size    & ModelNet40 \\ \hline

\hline

3DShapeNets \cite{wu20153d} & Voxelization  & $30^{3}$ & 77.32\% \\
OctNet \cite{riegler2017octnet} & Volumetric & $128^{3}$ &  86.5\% \\
\hline
MVCNN \cite{su2015multi} & Multi-view & $12\times224^{2}$ &  90.1\%  \\
GVCNN \cite{feng2018gvcnn} & Multi-view & 8 $\times$ & 93.1\% \\
\hline
DeepNet \cite{ravanbakhsh2016deep} & Points & $5000\times3$ &  90.0\% \\
ECC \cite{simonovsky2017dynamic} & Points & 1000 $\times$3 &   83.2\% \\
DGCNN \cite{wang2018dynamic}  & Points & $1024 \times 3$ &  92.2\% \\
Kd-Net \cite{klokov2017escape} & Points  & $2^{15} \times 3$ &  88.5\% \\
KPConv \cite{thomas2019kpconv} & Point & $7000 \times 3$ & 92.9\% \\
PointNet \cite{qi2017pointnet} & Points & $1024 \times 3$ &  89.2\%  \\
PointNet++ \cite{qi2017pointnet++} & Points & $1024 \times 3$ & 90.7\%  \\
3DmFV-Net \cite{Ben20183DmFV} & Points & 2048 $\times$ 3 &  91.4\% \\
FoldingNet \cite{yang2018foldingnet} & Points  & $2048\times3$ &  88.4\%  \\
KC-Net \cite{shen2018mining} & Points & $1024\times3$  &  91.0\% \\
PointCNN \cite{li2018pointcnn} & Points & $1024 \times 3$ &  92.5\% \\
PCNN \cite{atzmon2018point} & Points & $1024 \times 3$ &  92.3\% \\ 
RGCNN \cite{te2018rgcnn} & Points & $1024 \times $3&  90.5\%\\
ShapeContextNet \cite{xie2018attentional} & Points & $1024 \times 3$ &  90.0\%\\
Spec-GCN \cite{wang2018local} & Points & $2048 \times 3$ &  92.1\%\\
SRN \cite{duan2019structural} & Points & $1024\times3$ &  91.5\% \\
Point2Node \cite{han2019point2node} & Points & $1000 \times 3$ &  93.0\%\\
Point2Sequence \cite{liu2019point2sequence} & Points & $1024\times3$ & 92.6\%   \\
PointConv \cite{wu2019pointconv} & Points & $1024 \times 3$ & 92.5\%\\
$\Psi$-CNN \cite{lei2019octree} & Points & - &  92.0\%\\ 
FPConv \cite{lin2020fpconv} & - & - & 92.5\% \\
Point Transformer \cite{zhao2020point} & Points & $1024 \times 3$ & 89.6\%\\
SPH3D-GCN \cite{lei2020spherical} & Points & $1000 \times 3$ &  92.1\% \\
DR-Net \cite{qiu2021dense} & Points & $1024 \times 3$ & 93.1\% \\
Point Transformer \cite{engel2020point} & Points &  $1024 \times 3$ & 92.8\% \\ 
PT \cite{zhao2020point} & Points &  $1024 \times 3$ & 92.9\% \\
PCT \cite{guo2020pct} & Points & $1024 \times 3$ & 93.2\%  \\
PRA-Net \cite{cheng2021net} & Points & $1024 \times 3$ & 93.2\% \\
PAConv \cite{xu2021paconv} & Points & $1024 \times 3$ & 93.2\%\\
\hline
\textbf{Ours} & Points & $1024 \times 3$ & \textbf{93.2}\%\\

\hline
PointNet++ \cite{qi2017pointnet++}& Points+normals & $5000\times6$ &  91.9\%   \\
SpiderCNN \cite{xu2018spidercnn} & Points+normals & $1024 \times 6$ &  92.4\% \\
SFCNN \cite{rao2019spherical} & Points+normals & $1024\times6$ &  92.3\% \\
PointWeb \cite{zhao2019pointweb} & Points+normals & $1024 \times 6$ &  92.3\% \\
ELM \cite{fujiwara2020neural} & Points+normals & $2048 \times 6$& 93.2\% \\
FPConv \cite{lin2020fpconv} & Points+normals & - &  92.5\% \\
\hline
\end{tabular}
\label{tab:cls}
\end{table}

\begin{figure*}
\centering
\includegraphics[width=5in]{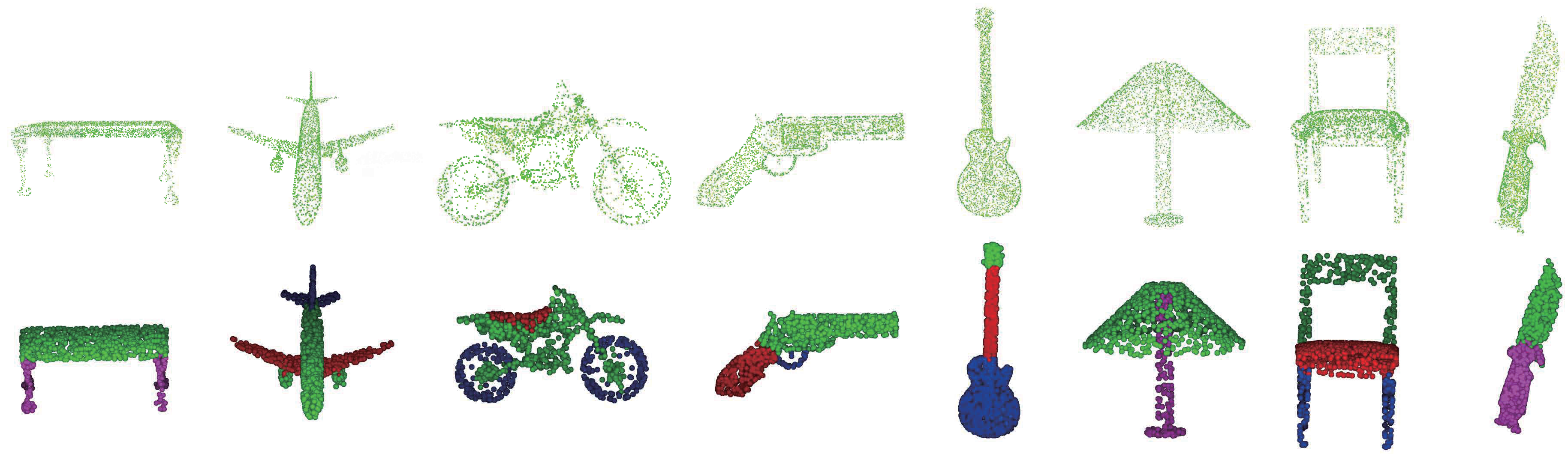}
\caption{Visualization of part segmentation on ShapeNet Part.}
\label{fig:partsegmentaion}
\end{figure*}

\begin{table*}[htbp]
\caption{Part segmentation results on ShapeNet part dataset. The mean IoU across all the shape instances and IoU for each category are reported. }
\label{PartSegmentation}
\centering
\resizebox{\textwidth}{!}{
\begin{tabular}{c|c|cccccccccccccccc}
\hline
Method & mIoU & aero & bag & cap & car & chair & ep & guitar & knife & lamp & laptop & motor & mug & pistol & rocket & skate & table\\
\hline
ShapeNet \cite{yi2016scalable} & 81.4 & 81.0 & 78.4 & 77.7 & 75.7 & 87.6 & 61.9 & 92.0 & 85.4 & 82.5 & 95.7 & 70.6 & 91.9 & \textbf{85.9} & 53.1 & 69.8 & 75.3 \\
PointNet \cite{qi2017pointnet} & 83.7 & 83.4 & 78.7 & 82.5 & 74.9 & 89.6 & 73.0 & 91.5 & 85.9 & 80.8 & 95.3 & 65.2 & 93.0 & 81.2 & 57.9 & 72.8 & 80.6 \\
PointNet++ \cite{qi2017pointnet++} & 85.1 & 82.4 & 79.0 & 87.7 & 77.3 & 90.8 & 71.8 & 91.0 & 85.9 & 83.7 & 95.3 & 71.6 & 94.1 & 81.3 & 58.7 & 76.4 & 82.6\\
KD-Net \cite{klokov2017escape} & 82.3 & 80.1 & 74.6 & 74.3 & 70.3 & 88.6 & 73.5 & 90.2 & 87.2 & 71.0 & 94.9 & 57.4 & 86.7 & 78.1 & 51.8 & 69.9 & 80.3 \\
SO-Net \cite{li2018so} & 84.9 & 82.8 & 77.8 & 88.0 & 77.3 & 90.6 & 73.5 & 90.7 & 83.9 & 82.8 & 94.8 & 69.1 & 94.2 & 80.9 & 53.1 & 72.9 & 83.0 \\
RGCNN \cite{te2018rgcnn} & 84.3 & 80.2 & 82.8 & \textbf{92.6} & 75.3 & 89.2 & 73.7 & 91.3 & 88.4 & 83.3 & 96.0 & 63.9 & \textbf{95.7} & 60.9 & 44.6 & 72.9 & 80.4 \\
DGCNN \cite{wang2018dynamic} & 85.2 & 84.0 & 83.4 & 86.7 & 77.8 & 90.6 & 74.7 & 91.2 & 87.5 & 82.8 & 95.7 & 66.3 & 94.9 & 81.1 & \textbf{63.5} & 74.5 & 82.6\\
SRN \cite{duan2019structural} & 85.3 & 82.4 & 79.8 & 88.1 & 77.9 & 90.7 & 69.6 & 90.9 & 86.3 & 84.0 & 95.4 & 72.2 & 94.9 & 81.3 & 62.1 & 75.9 & 83.2 \\
SFCNN \cite{rao2019spherical} & 85.4 & 83.0 & 83.4 & 87.0 & \textbf{80.2} & 90.1 & 75.9 & 91.1 & 86.2 & 84.2 & \textbf{96.7} & 69.5 & 94.8 & 82.5 & 59.9 & 75.1 & 82.9 \\
PointConv \cite{wu2019pointconv}& 85.7 & - & - & - & - & - & - & - & - & - & - & - & - & - & - & -\\
3D-GCN \cite{lin2020convolution} & 85.1 & 83.1 & 84.0 & 86.6 & 77.5 & 90.3 & 74.1 & 90.9 & 86.4 & 83.8 & 95.6 & 66.8 & 94.8 & 81.3 & 59.6 & 75.7 & 82.6 \\ 
ELM \cite{fujiwara2020neural} & 85.2 & 84.0 & 80.4 & 88.0 & 80.2 & 90.7 & \textbf{77.5} & 91.2 & 86.4 & 82.6 & 95.5 & 70.0 & 93.9 & 84.1 & 55.6 & 75.6 & 82.1 \\ 
Weak Sup. \cite{xu2020weakly} & 85.0 & 83.1 & 82.6 & 80.8 & 77.7 & 90.4 & 77.3 & 90.9 & 87.6 & 82.9& 95.8 & 64.7 & 93.9 &79.8 & 61.9& 74.9 & 82.9 \\
GBN \cite{9410405} & 85.9 & \textbf{84.5} & 82.2 & 86.8 & 78.9 & 91.1 & 74.5 & 91.4 & \textbf{89.0} & 84.5 & 95.9 & 69.6 & 94.2 & 83.4 & 57.8 & 75.5 & 83.5 \\
Point Transformer \cite{engel2020point} & 85.3 & - & - & - & - & - & - & - & - & - & - & - & - & - & - & -\\
PCT \cite{guo2020pct} & 85.8 & \textbf{84.5} & 83.5 & 85.9 & 78.7 & 90.9 & 75.1 & \textbf{92.1} & 87.0 & \textbf{85.0} & 95.9 & 69.6 & 94.5 & 82.2 & 61.4 & 76.0 & 83.0 \\
\hline
Ours & \textbf{86.0} & 83.6 & \textbf{84.7} & 86.3 & 79.8 & \textbf{91.1} & 71.2 & 90.2 & 88.6  & 84.9 & 95.9 & \textbf{72.8} &94.8 & 83.4 & 56.2 & \textbf{76.7} & 82.6 \\

\hline
\end{tabular}}
\end{table*}

\subsection{Part Segmentation}
Part segmentation is a challenging task aiming to assign a part label to each point in a given 3D point cloud object. 

\subsubsection{Datasets}
We evaluate the part segmentation task on the broadly used ShapeNet part dataset \cite{yi2016scalable}, which covers 16,881 shapes of 3D point cloud objects from 16 different categories. The object in each category are labeled with less than 6 parts, amounting to 50 parts in total, where each point is associated with a part label. These models are split into 14,007 examples for training and 2,874 models for testing. Here, we sample 2,048 points from the dataset. During training, we adopt the same data augmentation strategy as that for classification. We train our segmentation model 200 epochs with a mini batch size of 16. 

\subsubsection{Evaluation metric} The Intersection-over-Union (IoU) on points is considered as the metric for quantitatively evaluating the segmentation results of our model and comparison with other existing methods. Following the previous works \cite{qi2017pointnet}, we define the IoU of each category as the average of IoUs for all the shapes belonging to each category. In addition, the overall mean IoU (mIoU) is finally calculated by taking average of IoUs across all the shape instances. 

\subsubsection{Results} 

Table \ref{PartSegmentation} summarizes the performance comparison between our MLMST with several baselines. From the quantitative results, it can be clearly seen that our Transformer model reaches a much better part segmentation performance with the instance mIoU of \textbf{86.0\%}, outperforming the state-of-the-art approaches. The visualization of part segmentation on the ShapeNet part is given in Figure \ref{fig:partsegmentaion}. These results show the robustness of our MLMST to diverse shapes.




\subsection{Ablation Study}
In this section, we perform extensive ablation study to investigate the effectiveness of each individual components of our MLMST architecture using ModelNet40 for evaluation. Specifically, we adopt the single-resolution MLP as our baseline. Table \ref{tab:ablation} summarises the classification accuracy of different design choices. From these results, we can claim that the integration of PPT, MLT and MST modules achieve significant performance improvement over baseline. This further demonstrates that feature interactions across different levels and scales 
are beneficial to discriminative point cloud representation learning.

\begin{table}[]
\centering
\caption{Ablation analysis on our Multi-level Multi-scale Transformer architecture. }
\begin{tabular}{c|c} \hline
Method & Accuracy \\ \hline
Baseline & 92.2 \\
Baseline + PPT & 92.4\%\\
Baseline + PPT + MLT & 93.0\%\\
Baseline + PPT + MST & 92.8\%\\
MLMST & 93.2\% \\
\hline
\end{tabular}

\label{tab:ablation}
\end{table}

\subsection{Conclusion}

In this paper, we introduced, Mult-level Multi-scale Point Transformer, an end-to-end architecture relying on self-attention mechanism for point cloud analysis, which integrates three fundamental building modules, a point pyramid transformer, a multi-level transformer and a multi-scale transformer, to enrich contextual interaction across different levels and scales. Extensive experiments conducted on challenging benchmarks have demonstrated our MLMSPT achieves the state-of-the-arts performance on 3D object classification and segmentation. We believe that Transformer can play an important role in learning point cloud representation, therefore, further investigation of its development and application to various point cloud based tasks should be explored in future.

\section*{ACKNOWLEDGMENT}
This research was supported by the National Natural Science Foundation of China (No. 62002299), and the Natural Science Foundation of Chongqing of China (No. cstc2020jcyj-msxmX0126), and the Fundamental Research Funds for the Central Universities (No. SWU120005)


\bibliography{sn-bibliography}


\end{document}